\documentclass[letterpaper, 10 pt, conference]{ieeeconf}  

\IEEEoverridecommandlockouts                              % This command is only needed if 
\usepackage{graphics} % for pdf, bitmapped graphics files
\usepackage{times} % assumes new font selection scheme installed

\usepackage{booktabs}
\usepackage{multirow}
\usepackage{multicol}
\usepackage{colortbl}
\usepackage{adjustbox}
\usepackage{array}
\usepackage{colortbl}

\usepackage[dvipsnames]{xcolor}

\newcommand{\Tab}[1]{Tab.~\ref{tab:#1}}
\newcommand{\Sec}[1]{Sec.~\ref{sec:#1}}

\newcolumntype{G}[2]{%
    >{\columncolor[gray]{0.8}\adjustbox{angle=#1,lap=\width-(#2)}\bgroup}%
    l%
    <{\egroup}%
}
\newcolumntype{W}[2]{%
    >{\adjustbox{angle=#1,lap=\width-(#2)}\bgroup}%
    l%
    <{\egroup}%
}
\newcommand*\rot{\multicolumn{1}{W{38}{1em}}}% no optional argument here, please!
\title{\LARGE \bf Artificial Intelligence for Long-Term Robot Autonomy: A Survey}

\author{Lars Kunze$^{1}$, Nick Hawes$^{1}$, Tom Duckett$^{2}$, Marc Hanheide$^{2}$, Tom{\' a}{\v s} Krajn{\' i}k$^{3}$% <-this % stops a space
\thanks{$^{1}$Oxford Robotics Institute, Dept. of Engineering Science, University of Oxford, United Kingdom;
        {\tt\small \{lars|nickh\}@robots.ox.ac.uk}}%
\thanks{$^{2}$School of Computer Science, University of Lincoln, United Kingdom;
        {\tt\small \{tduckett|mhanheide\}@lincoln.ac.uk}}%
\thanks{$^{3}$Dept. of Computer Science, Faculty of Electrical Engineering, Czech Technical University in Prague;
        {\tt\small tomas.krajnik@fel.cvut.cz}}%
	\thanks{The work has been supported by EPSRC grant EP/R02572X/1, EU project No. 732737 (ILIAD) and CZ projects 17-27006Y and CZ.02.1.01/0.0/0.0/16\_019/0000765.}
}

\usepackage{tikz}
\usepackage{textcomp}

\newcommand\copyrighttext{%
  \footnotesize \textcopyright 2018 IEEE. Personal use of this material is permitted. Permission from IEEE must be obtained for all other uses, in any current or future media, including reprinting/republishing this material for advertising or promotional purposes, creating new collective works, for resale or redistribution to servers or lists, or reuse of any copyrighted component of this work in other works.}

\usepackage{eso-pic}

\begin{document}
% copyright notice for preprint
\AddToShipoutPicture*{\footnotesize \raisebox{1.1cm}{\hspace{1.8cm}\fbox{\parbox{\dimexpr\textwidth-\fboxsep-\fboxrule\relax}{\copyrighttext}}}}

\maketitle
\thispagestyle{empty}
\pagestyle{empty}

%%%%%%%%%%%%%%%%%%%%%%%%%%%%%%%%%%%%%%%%%%%%%%%%%%%%%%%%%%%%%%%%%%%%%%%%%%%%%%%%
\begin{abstract}

Autonomous systems will play an essential role in many applications across diverse domains including space, marine, air, field, road, and service robotics. 
They will assist us in our daily routines 
and perform dangerous, dirty and dull tasks. 
However, enabling robotic systems to perform autonomously in complex, real-world scenarios over extended time periods (i.e. weeks, months, or years) poses many challenges. 
Some of these have been investigated by sub-disciplines of Artificial Intelligence (AI) including navigation \& mapping, perception, knowledge representation \& reasoning, planning, interaction, and learning. 
The different sub-disciplines have developed techniques that, when re-integrated within an autonomous system, can enable robots to operate effectively in complex, long-term scenarios.   
In this paper, we survey and discuss AI techniques as `enablers' for long-term robot autonomy, current progress in integrating these techniques within long-running robotic systems, and the future challenges and opportunities for AI in long-term autonomy.
\end{abstract}

%%%%%%%%%%%%%%%%%%%%%%%%%%%%%%%%%%%%%%%%%%%%%%%%%%%%%%%%%%%%%%%%%%%%%%%%%%%%%%%%
\section{INTRODUCTION}
\label{sec:introduction}

Robot technology has improved tremendously  over the last decade. Consequently, autonomous robot systems have been able to operate 
in increasingly complex environments and 
for increasingly long periods of time, i.e. weeks, months, or years. When a fully modelled robot is deployed in a completely known, static environment, the challenge of long-term autonomy (LTA) reduces to one of robustness, i.e.~enabling the robot to remain operational for as long as possible. Without these simplifying assumptions autonomous robots face a number of interrelated challenges. 
We roughly characterise these challenges on two dimensions. The first refers to the application requirements, e.g., the robot platform (hardware and software), environment and tasks to be performed. The second dimension describes the long-term nature of these elements, e.g., if and how they change over time, whether their long-term nature can be fully characterised in advance (structured vs. unstructured), and how observable they are.
For example, in many long-term applications, the environment will change over the lifetime of the system. These changes could be short-term (e.g.\ things moving within the robot's field of view), medium-term (e.g.\ furniture moving between visits to a room, parked cars changing positions on roads), or long-term (e.g.\ seasonal changes, plant growth, wear to surfaces). 
In addition, parts of the environment may not be fully known before deployment or new objects may appear. 
In AI terms this means dealing with an \emph{open world}.
It is also possible that the end-user may change the tasks or how the robot should accomplish them, 
or the robot itself may need to adapt to new tools, techniques (e.g.\ AI algorithms) or knowledge as they become available.

In this paper, we survey systems and approaches that address the challenges of LTA using techniques from AI. 
We focus on both AI techniques used by robot systems deployed for extended periods in real-world environments, i.e.\ no highly contrived or controlled settings, and no single run examples (\Sec{domains}),
and
techniques which may have yet to be convincingly demonstrated in a real-world long-term 
application, i.e.\ only on a data set or in a lab (\Sec{areas}), but align well with future needs of long-term autonomous systems.
We further discuss the future challenges and opportunities 
for AI in LTA (\Sec{challenges}).

By focussing on the above challenges 
we purposely exclude other applications where robots operate for long periods, but in relatively static, known settings. Specifically this means we do not cover current systems in manufacturing or intra-logistics. In both cases the majority of deployed robot systems have demonstrated significant longevity, but this is typically achieved through the creation of environments which are fully known and the dynamics are largely under control of the autonomous systems. Whilst this does not eliminate the need for AI techniques (e.g.\ long-term localisation~\cite{Gadd2015} and planning~\cite{kiva} for warehouse AGVs), it limits 
the LTA-specific challenges (e.g. environmental dynamics, lack of structure, open-endedness) seen in other domains.

Since autonomous systems research has a long history, special issues and surveys already exist that relate to LTA. 
For example, \cite{RAJAN20171,INGRAND201710} cover 
AI methods within integrated robot systems.
In perception, existing collections cover localisation and mapping in dynamic environments~\cite{doi:10.1177/0278364913511182,cadena2016}, calibration~\cite{ROB:ROB21619} and visual place recognition~\cite{lowry2016}.
However, this is the first survey that focuses specifically on AI techniques for enabling long-term robot autonomy.

\section{DOMAINS}
\label{sec:domains}

\begin{table*}%[t]
  \vspace{1em}
  \begin{center}
  {\small
\caption{Surveyed AI-enabled long-term autonomy robot systems.}
      \label{tab:surveyed-systems}
    \setlength{\aboverulesep}{0pt}
    \setlength{\belowrulesep}{0pt}
    \setlength{\extrarowheight}{.75ex}
    \begin{tabular}{p{3.3em}ccccccccp{3.5em}ccccccl}
      \toprule
      Domain & \multicolumn{8}{c}{Application Features} &  Duration &  \multicolumn{6}{c}{AI Areas} & System\\

      \cmidrule(lr){1-1}
      \cmidrule(lr){2-9}
      \cmidrule(lr){10-10}
      \cmidrule(lr){11-16}
      \cmidrule(lr){17-17}
      \\

      & \rot{Environment Variability}  
      & \rot{Task Diversity}  
      & \rot{Semantics}  
      & \rot{Dynamics} 
      & \rot{Partial Observability}
      & \rot{Cost \& Criticality}
      & \rot{Interaction \& Cooperation}
      & \rot{Level of Autonomy} 

      &
      & \rot{Navigation \& Mapping} 
      & \rot{Perception} 
      & \rot{KR \& Reasoning}
      & \rot{Planning} 
      & \rot{Interaction}
      & \rot{Learning} \\
      \midrule

     \multirow{2}{*}{Space} & \multirow{2}{*}{\textcolor{gray!50}{L}} & \multirow{2}{*}{\textcolor{gray!50}{L}} & \multirow{2}{*}{\textcolor{gray!50}{L}} &   \multirow{2}{*}{\textcolor{gray!50}{L}} & \multirow{2}{*}{H} & \multirow{2}{*}{H} & \multirow{2}{*}{\textcolor{gray!50}{L}} & \multirow{2}{*}{\textcolor{gray!100}{M}} & Years & $\circ$ & $\bullet$ &  $-$  & $\bullet$ & $\circ$ & $-$ & Opportunity~\cite{mapgen,rover_nav}\\\cmidrule{10-17}

      &  &  &  &  &  & & & & Years  & $\circ$ & $\bullet$ & $-$  &  $\bullet$ & $\circ$ & $-$ & IPEX~\cite{ipex}\\\hline

     \multirow{2}{*}{Marine}& \multirow{2}{*}{\textcolor{gray!100}{\textcolor{gray!100}{M}}} & \multirow{2}{*}{\textcolor{gray!50}{L}} & \multirow{2}{*}{\textcolor{gray!50}{L}} &  \multirow{2}{*}{\textcolor{gray!100}{M}} & \multirow{2}{*}{H} & \multirow{2}{*}{H} & \multirow{2}{*}{\textcolor{gray!50}{L}} & \multirow{2}{*}{H} & Days & $\circ$ & $\bullet$ & $\circ$ & $\bullet$ & $-$ & $\circ$ & AUVs~\cite{ice,chrpa2015mixed}  \\\cmidrule{10-17}

     &  &  &  &  & & & & & Months & $\circ$ & $\circ$ & $-$ & $\circ$ & $-$ & $-$ & Gliders~\cite{gliders}  \\\hline

      Air   & \textcolor{gray!100}{M} & \textcolor{gray!100}{M} & \textcolor{gray!100}{M} &  H & H & H & \textcolor{gray!100}{M} & \textcolor{gray!100}{M} & Days  & $\circ$ & $\bullet$ &  $\circ$  & $\circ$ & $-$ & $-$ &AtlantikFlyer~\cite{oettershagen_2016} \\\hline

     \multirow{2}{*}{Field} & \multirow{2}{*}{H} & \multirow{2}{*}{\textcolor{gray!100}{M}} & \multirow{2}{*}{\textcolor{gray!50}{L}} &  \multirow{2}{*}{\textcolor{gray!100}{M}} & \multirow{2}{*}{H} & \multirow{2}{*}{\textcolor{gray!100}{M}} & \multirow{2}{*}{\textcolor{gray!100}{M}} & \multirow{2}{*}{\textcolor{gray!100}{M}} & Days & $\bullet$ & $\bullet$ & $\circ$ & $-$ & $\circ$ & $\circ$ & VT\&R2 \cite{paton_2018} \\\cmidrule{10-17}
      &  &  &  &   &  & & & & Years & $\bullet$ & $\bullet$ & $\circ$ & $-$ & $-$ & $\circ$ & BearNav \cite{krajnik_2010,krajnik_2017} \\\hline

     \multirow{4}{*}{Road} & \multirow{4}{*}{\textcolor{gray!100}{M}} & \multirow{4}{*}{\textcolor{gray!50}{L}} & \multirow{4}{*}{\textcolor{gray!100}{M}} &  \multirow{4}{*}{H} & \multirow{4}{*}{\textcolor{gray!100}{M}} & \multirow{4}{*}{H} & \multirow{4}{*}{\textcolor{gray!100}{M}}  & \multirow{4}{*}{\textcolor{gray!50}{L}}  & Days & $\circ$ & $\bullet$ & $\bullet$ & $\circ$ & $-$ & $\circ$ & VaMP~\cite{Dickmanns:2007:DVP:1205575}  \\ \cmidrule{10-17}

     &  &  &  &  & & & & & Days & $\circ$ & $\bullet$ & $\circ$ & $\circ$ & $-$ & $\circ$ & ARGO~\cite{broggi1999argo}  \\  \cmidrule{10-17}

     &  &  &  &  & & & & & Months & $\circ$ & $\bullet$ & $\circ$ & $\circ$ & $-$ & $\circ$ & PANS~\cite{pans1995}  \\\cmidrule{10-17}

     &  &  &  &  & & & & & Months & $\circ$ & $\bullet$ & $\circ$ & $\circ$ & $-$ & $\circ$ & VIAC~\cite{doi:10.1504/IJVAS.2012.051250} \\\hline

     \multirow{5}{*}{Service} & \multirow{5}{*}{H} & \multirow{5}{*}{H} & \multirow{5}{*}{H} &  \multirow{5}{*}{\textcolor{gray!50}{L}} & \multirow{5}{*}{H} & \multirow{5}{*}{\textcolor{gray!50}{L}}& \multirow{5}{*}{H}& \multirow{5}{*}{\textcolor{gray!100}{M}} & Days & $\bullet$ & $\circ$ & $\circ$ & $\bullet$ & $\bullet$ & $\circ$ & Rhino~\cite{Rhino1}  \\\cmidrule{10-17}

     & & & & & & & & & Days & $\bullet$ &$\circ$ & $\circ$ & $\bullet$ & $\bullet$ & $\circ$ & Minerva~\cite{Minerva}  \\\cmidrule{10-17}

     & & & & & & & & & Days & $\bullet$ & $\circ$ & $\circ$ & $\circ$ & $\bullet$ & $\circ$ & Willow Garage~\cite{Meeussen2011a}  \\\cmidrule{10-17}

     & & & &  & & & & & Months & $\bullet$ & $\bullet$ & $\bullet$ & $\bullet$ & $\bullet$ & $\bullet$ & STRANDS~\cite{Hawes2016}  \\\cmidrule{10-17}

      &  &  &  &  & & & & & Years & $\bullet$ & $\bullet$ & $\bullet$ & $\bullet$ & $\bullet$ & $\bullet$ & CoBot~\cite{cobots1000km}\\

        \bottomrule
    \end{tabular}
     \begin{center}
      Legend:  \textcolor{gray!50}{L} \emph{low}, \textcolor{gray!100}{M} \emph{medium}, H \emph{high}, $-$ \emph{not integrated}, $\circ$ \emph{partially integrated}, $\bullet$ \emph{fully integrated} 
    \end{center}
     %\hfill{}
    }
  \end{center}
  \vspace{-1em}
\end{table*}

Long-term autonomous robots have been deployed in a variety of domains including space, marine, air, field, road, and service. 
\Tab{surveyed-systems} provides an overview of these domains and selected systems characterised across common features. 
In this survey, we adopt the notation of \cite{INGRAND201710} and characterise domains by application features: \emph{environment variability}, \emph{task diversity}, \emph{semantics}, \emph{dynamics}, \emph{partial observability}, \emph{cost \& criticality}, \emph{interaction \& cooperation}, and \emph{level of autonomy}.
As in \cite{INGRAND201710}, all features are qualitatively assessed using three levels (\emph{low, medium, high}). Please see the aforementioned paper for a detailed discussion of the features.
In this work, we focus on the assessment of deployed robot systems. To this end, we assess them by the duration of their deployment (\emph{days, months, years}) and the level of integration of different AI areas (\emph{not, partially, or fully integrated}).

\paragraph*{Space}
\label{sec:space}

Due to extreme communication delays and limited prior access, effective extra-terrestrial exploration requires autonomous systems.
NASA's Opportunity rover has recently passed its 5,000th day operating on the surface of Mars. 
Its autonomous capabilities come from a mixed-initiative task planner, and an autonomous navigation system. 
The planner (MAPGEN,~\cite{mapgen}) is used to automatically create a daily mission schedule, which is then refined by terrestrial scientists. The navigation system uses stereo cameras to build 3D models for terrain traversability and path planning~\cite{rover_nav}. 
LTA has also been a growing part of satellite operations, e.g. the Intelligent Payload Experiment (IPEX) demonstrating over a year of autonomous information gathering using planning and image processing technology~\cite{ipex}. 

\paragraph*{Marine}
\label{sec:marine}

Due to the limits of communication through water, and the difficulties in fully mapping deployment environments, there are parallels between the requirements for, and benefits of, autonomy in marine and space robotics.
Autonomous wave gliders are routinely deployed for long durations, with missions measured in thousands of kilometres and hundreds of days (e.g. 7,400 km in 221 days~\cite{gliders}). 
Gliders are relatively simple, low-powered robots. More powerful systems have been deployed for days of autonomous operation, e.g. for navigation under ice~\cite{ice}. 
The benefits of AI planning have been shown in field trials~\cite{chrpa2015mixed} and controlled settings targeting LTA~\cite{palomeras2016toward}.

\paragraph*{Air}
\label{sec:air}
The fundamental factor that makes long-term operation of aerial systems difficult is energy.
The authors of~\cite{oettershagen_2016} argue that to achieve perpetual autonomous flight, the UAV has to plan its path according to global and local weather conditions, wind fields, and thermal updrafts.  
An alternative to perpetual flight is the ability to interrupt the flight to recharge, 
like the lake monitoring system in~\cite{peloquin_2017}.

\paragraph*{Field}
\label{sec:field}

Field robotics deals with unstructured and dynamic environments in diverse domains such as forestry, agriculture, mining, construction, etc.
Bechar and Vigneault~\cite{Bechar2016} characterise such domains according to the level of structure present in both the environment and the objects relevant to the robot.
The majority of current field robots utilise GPS-based auto-steer systems that follow pre-determined paths with otherwise limited use of AI capabilities.
An alternative approach uses visual `teach and repeat' to enable robust navigation in field environments. In these approaches systems are driven along a training route and then they repeat the route autonomously~\cite{Furgale2010}. 
Krajnik et al. [17] show that their teach-and-repeat method is robust to seasonal appearance changes. Paton et al. [16] showed that the integration of multiple experience-based representations [39] results in a system capable of long- term autonomous navigation despite drastic changes to the environment appearance.

\paragraph*{Road}
\label{sec:road}

The PANS platform \cite{pans1995} was one of the first autonomous vehicles that drove a long distance (6,000 miles, $98.2\%$ autonomous driving) on public roads over a period of six months using a vision-based driving system. 
It used a neural network  to learn a mapping between road images and appropriate vehicle turn radiuses from human demonstrations.
At the same time, the driverless car VaMP~\cite{Dickmanns:2007:DVP:1205575}
drove more than 1,000 miles ($95\%$ autonomous driving). 
The vision-based driving system of the ARGO project~\cite{broggi1999argo} achieved a similar result (1,200 miles, $94\%$ autonomous driving).  
Through learning it was able to adapt to new road conditions (lane width and lane position). 
More recently, several vehicles covered a distance of 13,000 km from Italy to China using a leader-follower approach in the VisLab Intercontinental Autonomous Challenge (VIAC) (2010) \cite{doi:10.1504/IJVAS.2012.051250}. 

\paragraph*{Service}
\label{sec:service}

We characterise service robots as robots that work for, or alongside, humans in environments that are not specially adapted for their presence.
Service robots must cope with: dynamic environments (due to people moving, day-night changes, etc.); open worlds (due to people); and changing task requirements.
Large-scale research initiatives have deployed mobile robot systems capable of LTA in museums (the seminal Rhino~\cite{Rhino1} and Minerva~\cite{Minerva}), offices (Willow Garage~\cite{Meeussen2011a}, CoBot~\cite{cobots1000km} and STRANDS~\cite{Hawes2016}), stores~\cite{Gross2009a}, and care environments (STRANDS~\cite{Hanheide2017} and Tangy~\cite{tangy}). All of these robots were deployed for at least multiple weeks, and most around na\"ive users. Most of these systems were deployed at intervals in the same environment (e.g. daily). The STRANDS and Willow Garage systems also attempted \emph{continuous} autonomous operation, managing a maximum of 28 and 13 days respectively of uninterrupted operation. These research systems have given rise to the current generation of autonomous service robots operating in human-populated spaces. Examples include Bossanova's stock checking robots in Walmart stores, Knightscope's security robots, and Savioke's robot hotel butlers.

\paragraph*{Conclusion}
\label{sec:domain:conclusion}

With respect to AI areas, \emph{Navigation \& Mapping} and \emph{Perception} are the only areas that were present in all surveyed systems. This is no surprise as they provide robots with very fundamental capabilities. \emph{KR \& Reasoning} as well as \emph{Planning} were both supported by most systems. However, we hypothesise that work on \emph{KR \& Reasoning} in space and marine is limited due to the lack of semantics in these domains. Furthermore, it is interesting to note that systems only partially (if at all) support \emph{Interaction} and \emph{Learning} in most domains (with an exception of the service domain). Although these areas are well researched in general, they haven not been extensively covered in long-term scenarios. Hence, we believe that there are many open challenges and research opportunities for both areas (and in their intersection) as we point out in \Sec{challenges}.

In all domains, LTA systems inherently present an \emph{integration} challenge, particularly when different AI abilities need to work together. 
Over the past years, there has been an increasing trend toward the (re-)integration of AI techniques within robotics. 
To cope with challenging environments and tasks, robots typically integrate: localisation and navigation; object and/or person perception; plus task planning and/or scheduling. 
However, although the integration of AI techniques at system-level is an essential part of all research projects, there is no standard solution and little research on how to combine modules from different areas of AI.

Robotic software development~\cite{software2009} and robotic middleware projects such as the Robot Operating System (ROS)~\cite{ros2009} provide researchers with common methods to integrate their software components and components of others in a structured way. Some frameworks build on top of such middlewares and integrate particular AI methods in the context of long-term navigation planning and task scheduling (STRANDS~\cite{Hawes2016}), planning and execution (ROSPlan~\cite{ICAPS1510619}), and knowledge-enabled perception (RoboSherlock~\cite{beetz15robosherlock}). In general, these frameworks make it easier to integrate and use different AI methods. However, overall, there is still a lack of understanding and research in the area of \emph{system-level integration}. Hence, we believe system-level integration of AI methods and their evaluation will continue to be a major challenge in autonomous systems research.

\section{AI AREAS}
\label{sec:areas}

In this section we discuss how different areas of AI can enable autonomous robot systems to 
perform in real-world environments over extended periods of time.
This includes navigation \& mapping, perception, knowledge representation \& reasoning, planning, interaction, and learning.

\subsection{NAVIGATION \& MAPPING}
\label{sec:navigation}
Navigation is 
an essential ability 
for purposeful movement by autonomous robots.
One approach uses visual `teach and repeat' to enable robust navigation in field environments, where the robot learns a map while being driven along a training route and then repeats the route autonomously~\cite{Furgale2010, krajnik_2010, paton_2018}, as discussed earlier in \Sec{field}.
Recent work~\cite{ROB:ROB21669} demonstrated 
over 140 km of autonomous driving with an autonomy rate of 99.6\%, including driving at night-time. 
 
Over the past 30 years there has been huge interest in autonomous learning of environment models by robots, especially the problem of simultaneous localisation and mapping (SLAM)~\cite{cadena2016}. However, most approaches assume a static world and do not consider long-term updating of robot maps to reflect environment changes.
Here we briefly characterise several complementary strategies %from recent literature 
to enable long-term mapping and localisation in changing environments, primarily using long-term data sets for their experiments.

\paragraph*{Multiple representations} 
\label{sec:multiple-rep}

Long-running robots need to consider environment mapping as a never-ending process, and thus make decisions on what to remember and what to forget. However, deleting information from a map is risky since an observed change may only be temporary and the environment may yet revert to a previously observed state. 
One approach is thus to maintain multiple environment representations~\cite{lowry2016}, then to select the most relevant model at the current time for localisation and planning.
Early work~\cite{biber_2005} developed dynamic maps that handle changes through use of robust statistics and multiple local maps at different timescales, where the map that best explains the current sensor data is used for localisation.
The short-term maps are updated online while the longer-term maps are adapted offline based on long-term experience.
Stachniss \& Burgard~\cite{stachniss_mobile_2005} cluster local grid maps created at different times and learn distinct configurations of these locations. 
A related approach for 
pose-graph SLAM~\cite{konolige_2009} maintains multiple view-based representations of mapped locations, while discarding obsolete views, thus limiting overall map size. 
Similarly, Churchill \& Newman~\cite{churchill_2013} propose to integrate similar observations at the same spatial locations into `experiences' which are then associated with a given place. For localisation, they select the experience that best matches the visual input of the robot. 
An alternative approach is to keep the data from all mapping sessions and integrate them offline into a single, high fidelity representation 
\cite{summary_2015}.

\paragraph*{Robustness to appearance change}

A parallel strategy attempts to select the representation which is most stable in time.
Valgren \& Lilienthal~\cite{Valgren2010} demonstrated the robustness of local image features for localisation across seasons.
SeqSLAM~\cite{Milford2012} attempts to match sequences of images rather than individual images, achieving robust place recognition across seasons.
A method for learning long-term stable features is described by Dayoub et al.~\cite{dayoub_2011}, where image features detected across  mapping sessions are first stored in a short-term memory, which is used to filter out spurious observations, before being admitted to long-term memory.
A further approach involves learning to predict appearance changes across seasons~\cite{Neubert2013}, by learning the expected translation between a vocabulary of superpixels for different seasons and using this to generate predicted images for localisation at run-time.
Recent research showed that season-specific images can also be predicted using generative adversarial networks~\cite{cnn_horia,cnn_latif}.
Lowry \& Milford~\cite{lowry_2016} compare a similar appearance prediction technique with a change removal method and conclude that change removal is more robust and less data-intensive to train. 
Related work on laser-based localisation~\cite{Withers2017} uses long-term experience to learn error distributions for individual points in 3D point-cloud maps, which are then used during localisation to suppress the observations corresponding to map points with high errors.

\paragraph*{Learning about dynamics}

While the above approaches are mainly concerned with learning the persistent elements of the scene, 
another strategy attempts to model the dynamics.
Tipaldi et al.~\cite{Tipaldi2013}
use dynamic occupancy grids, which model the occupancy of each cell as a two-state Markov process, and showed that their approach improves localisation robustness in a car park environment. 
Kucner et al.~\cite{Kucner2013} learn conditional probabilities of neighbouring cells
in an occupancy grid to model typical motion patterns in
dynamic environments.
Krajnik et al.~\cite{Krajnik2017} 
proposed to represent rhythmic or periodic processes in the environment using Fourier analysis, and showed that the resulting spectral models obtained from long-term experience enable prediction of future environment states, improving localisation and navigation in human-populated environments.

Notable applications of long-term mapping include
visual survey of natural environments by 
an autonomous surface vessel surveyed a lake shore over a 14-month period~\cite{Griffith2016},
and a 4D reconstruction approach to crop monitoring over time~\cite{Dong2017}. The latter comprises a 3D SLAM pipeline, data association to find correspondences between crop rows and sessions, and optimisation of the full 4D reconstruction.

Finally, complementary work on topological and semantic mapping may further enhance long-term robustness to change, by abstracting away from the finer details of metric and feature-based representations, although a detailed review is beyond the scope of this paper.
Current trends suggest that future work on long-term navigation and mapping will include more application-specific developments across all domains, as long-running systems continue to be deployed in practice, and development of richer environment representations including especially more semantics and integration of more perceptual and contextual cues.

\subsection{PERCEPTION}
\label{sec:perception}

In addition to perception algorithms for navigation and mapping, 
autonomous robots need general perception routines for object recognition and scene understanding.
Indeed, early approaches to mapping of dynamic environments were object-centric.
These methods identify moving objects and remove them from the maps~\cite{Wolf2005}
or use them as moving landmarks for self-localisation~\cite{Wang2007}. However, not all dynamic objects actually move at the moment of mapping, meaning 
that their identification requires long-term observations.

To address this challenge, Ambrus et al.~\cite{Ambrus2014} processed 
several 3D point clouds of the same environment recorded over several weeks to identify and separate movable objects, and refine the static environment structure at the same time.
Biswas \& Veloso~\cite{Biswas2014} proposed an approach for long-term localisation based on explicit reasoning about object categories including mapped objects, unmapped static objects and unmapped dynamic objects.
Bore et al.~\cite{Bore2018} detect and localise objects in large environments, where objects can change locations between observations by the robot, 
while assuming a closed world to ensure computational tractability.
%​

Other approaches enable open-ended learning of new object categories during long-term operation, e.g.\
using spatial context information to query possible category labels from semantic knowledge on the web.
Recent work includes 
an embodied system for open-ended learning and manipulation of new object categories, based on human-robot interaction~\cite{HamidrezaKasaei2018},
and
a lifelong learning framework in which a human user can direct  a robot to capture domain-relevant data for training classifiers of household objects~\cite{Eriksen2018}.

Future service robots would also benefit from techniques to improve their perception of people over time, e.g.\ by integrating long-term experience in tracking-learning-detection~\cite{Kalal2012}
and tracking-learning-classification~\cite{yan2017online} approaches,
and learning the long-term activity patterns of people~\cite{Molina2018}.
This would in turn enable robots to adapt to and move more harmoniously with the expected flow of humans.
Long-term applications involving interaction with specific people
also require algorithms for person re-identification 
at different temporal scales (very short term,
same day, different day).
For these cases, different assumptions can be made based on the persistence
of supporting cues (e.g., position at dinner table, clothing, size/stature, hair colour, facial features).
Recent work integrates person re-identification with multi-target multi-camera tracking~\cite{BeyerBreuers2017Arxiv}, however, adaptation of person-specific appearance models over long time-periods remains an open challenge for autonomous robots.
Direct parallels may be drawn for other related challenges such as recognition of human activities, where long-term experience can be leveraged to improve performance over time~\cite{lirolem23297}.
In general, 
most prior work on perception considers only 
the initial training phase prior to deployment of the robot, but not the 
ongoing adaptation of the learned models during long-term operation.

\subsection{KNOWLEDGE REPRESENTATION \& REASONING}
\label{sec:represenation}

Knowledge representation (KR) is concerned with representing 
the world (in particular domains with rich semantics, see \Tab{surveyed-systems}), 
and is closely linked to many other AI areas including perception, planning, and learning.  
The previous sections already discussed important aspects of representations in the context of navigation and perception.

In general, KR goes hand and hand with reasoning as both decision making and inference are tightly coupled with the way the knowledge is represented. Long-term, autonomous robot systems that are deployed in a real-world environments require KR and reasoning capabilities to represent various aspects of the world and reason about them, in particular when they change over time. Therefore, AI areas such as spatio-temporal reasoning, non-monotonic reasoning, and belief revision are of great importance in long-term scenarios. 

Several works investigated models that infer locations of entities in space and time. 
Mason et al. \cite{mason2012} proposed an object-based semantic world model for long-term change detection in dynamic environments, and 
\cite{toris2017} modelled the temporal persistence of objects. 
Similarly, Krajnik et al. \cite{krajnik2015} proposed frequency-based spatio-temporal models for reasoning about the location of people. 
Such spatio-temporal information is essential in long-term scenarios as it can inform AI planners (cf. \Sec{planning}) about non-stationary costs and/or rewards. 
Santos et al. \cite{santos2016ral} presented a first lifelong information-driven approach to spatio-temporal exploration that incrementally completes and refines environment maps. 

In LTA it is of great importance that robots can access and learn from their own experience. 
OPEN-EASE \cite{tenorth15openease} is a KR infrastructure that makes experience data from robots and human manipulation episodes semantically accessible. 
Users can retrieve experiences and query what the robot perceived, reasoned and did. Balint-Benczedi et al.~\cite{balintbe17icar} propose a more specialised framework for storing and retrieving perceptual memories for long-term manipulation tasks. 
Similarly, \cite{doi:10.1080/0952813X.2015.1134679} propose a long-term knowledge acquisition framework using contextual information in a memory-inspired robot architecture. The framework allows robots to memorise their perceptions and to recall them, e.g.\ in a manipulation task.

To cope with the challenges of open worlds, novelty and anomaly detection is of great importance. To this end, \cite{Winder:2016:FAR:3061053.3061210} proposed a framework for anomaly reasoning which includes the recognition and interpretation of unfamiliar and familiar objects appearing in unexpected contexts. This aspect of KR and reasoning is  strongly linked to work in adaptation and learning (\Sec{learning}) as it can trigger  learning in LTA systems.

\subsection{PLANNING}
\label{sec:planning}

AI planning and scheduling technologies, which determine the sequence of actions necessary to achieve a task, are often used to adapt the robot's behaviour online to account for environment or task dynamics~\cite{INGRAND201710}. We have seen planning systems deployed on almost all LTA systems. For example, planning approaches were used to produce daily task lists and the associated action sequences for the Opportunity rover~\cite{mapgen}, and the STRANDS~\cite{Hawes2016}, CoBot~\cite{Veloso:2012} and Tangy~\cite{tangy} service robots, allowing these robots to adapt their behaviour to the needs of their users. AUVs used planning to deal with changing environmental conditions and resources~\cite{chrpa2015mixed}, while logistics systems used planning to enable large numbers of robots to cope with variety in customer orders~\cite{kiva}.

Planning approaches vary in their ability to represent critical elements of 
a system's long-term experience. The aforementioned systems vary in terms of whether or not they model the effects/dependencies of a robot's actions on time or resources (such as battery), or under uncertainty. They also vary in how they handle oversubscription (choosing between multiple goals, a key issue in integrating exploration). In general, planning algorithms in LTA robots are embedded in a wider integrated system which handles the parts omitted from the planning model (e.g. replanning on failed on actions, or reactively triggering charging on low battery, or managing goal choices). More generally, an executive control system which manages tasks, and responds to opportunities and failures, is an essential part of a robot architecture for long-term autonomy~\cite{palomeras2016toward,chrpa2015mixed,Hawes2016,Marder-Eppstein2010,Rhino1}. Such a system prevents the robot getting stuck in behavioural loops, and provides recovery mechanisms to address autonomy-hindering failures. This behaviour can be seen in simple yet effective form in the finite state controller of Willow Garage's office marathon system~\cite{Marder-Eppstein2010}, through the planner-based executives of Rhino~\cite{Rhino1} and Minerva~\cite{Minerva}, through to the T-Rex executive used on fielded AUV teams~\cite{chrpa2015mixed}.

The planned behaviour of the STRANDS~\cite{Hawes2016} and CoBot~\cite{Veloso:2012} robots was generated using models learnt during execution: STRANDS robots created optimal task and navigation plans from learnt MDP models of environmental dynamics~\cite{LPH14b,jpulido2015NowOrLater}; CoBot robots learnt and planned with models which predicted when humans would be available to help complete a task~\cite{Rosenthal2012}. These robots were therefore able to adapt their task and navigation plans 
over the long term.

As robots become more adept at navigation and manipulation in less structured environments, we expect to see planning playing an even more prominent role in logistics, since a greater degree of variation will need to be managed autonomously over the life of the robot. We also expect to see overlap in learning and planning/optimisation processes in autonomous transport systems~\cite{miller2017predictive}, where system performance will need to be managed over  variation in demand and execution resources.  
Current trends also include the augmentation of 
plans or policies created by mission or task planning systems 
with richer execution knowledge~\cite{iocchi2016practical}. This hybrid approach allows mission planning to address long-term or large-scale problems with abstract, computationally tractable models, but at execution time behaviour is guided by richer models which allow  appropriate responses to dynamic events.

\subsection{INTERACTION}
\label{sec:interaction}

Some of the most challenging application domains for long-running robots 
involve interacting autonomously with a diverse range of users, offering opportunities for the robots not only to learn and adapt from this experience, but also facilitating longitudinal studies to gain a better understanding of long-term engagement of humans with autonomous robots.
In general, humans 
and other actors
introduce a level of dynamics and non-predictability into any application scenario, and hence pose dedicated challenges for LTA systems, also indicated in 
\Tab{surveyed-systems},
where \emph{environment variability} is considered high in domains with a high level of \emph{interaction and cooperation}.  
However, long-term Human-Robot Interaction (HRI) studies with truly autonomous \emph{social} robots are still a rarity today, as many researchers resort to Wizard-of-Oz settings~\cite{Riek2012a}, where subjects in studies are deceived into believing a robot is acting autonomously while it is in fact remote-controlled by a human operator. 
Among the most explored domains for long-term autonomous systems with an emphasis on 
interaction are museums~\cite{Minerva,Nourbakhsh2003}, care~\cite{Hanheide2017}, domestic~\cite{Graaf2017}, retail~\cite{Gross2009a,Foster2016}, hospitality~\cite{Pinillos2016}, and educational environments~\cite{Kanda2007, Jacq2016,Baxter2017}.

A recent survey~\cite{Leite2013a} identified key domains for long-term interactive robotic systems including `Health Care and Therapy', `Education', `Work Environments and Public Spaces', and `At Home', discussing a total of 45 different studies in these fields. 
From their analysis, the key conclusion drawn regarding autonomy is the lack of but also need for more learning and adaptation. Indeed several systems mentioned above (e.g.\ \cite{Foster2016, Baxter2017}) develop personalised models to maintain an interaction context. 
Such individualised user profiles are one of the key abilities required to enable interaction in long-running autonomous systems~\cite{Rossi2017}. 

Hence, interaction in the context of long-term autonomy must not only be seen as a challenge, but also as an opportunity, where representations can be learned or adapted in an \emph{in-situ} fashion to improve a system's autonomous behaviour from exploiting long-term interactions with users. \cite{Rosenthal2012} propose a model enabling the robot to predict when humans are most likely available to help a robot, while \cite{Hanheide2017} follow similar ideas, learning spatio-temporal usage patterns to maximise the utility of the mobile robot.

\subsection{LEARNING}
\label{sec:learning}

Machine learning plays a role in many of the above areas,
and is clearly a key enabling technology for LTA. Beyond this component role, a cluster of learning types are specifically suited to LTA. In general we see techniques that allow a robot to learn \emph{during operation} (rather than during a design phase) as crucial to success in LTA applications. 
%This is due to an assumption that a long-term deployment (at least in open/dynamic worlds) will mean 
Long-term deployment in open/dynamic worlds means that any knowledge or experience the robot starts with is unlikely to be sufficient to cover the behaviour required of it during operation. Thus learning during operation is essential to delivering good performance. We have seen this from 
%lifelong mapping approaches, through
the relatively low level of 
estimating cost and probability models for planning~\cite{Hawes2016}, up to learning new object~\cite{young:hal-01370140} and activity models for service robotics tasks~\cite{Duckworth:2016}. Since it is hard to receive supervision signals during long-term autonomous operation, the majority of the online learning techniques employed by LTA systems are unsupervised. 

By definition, a robot is restricted to a fixed viewpoint in space and time. This means that it is limited in the experiences, and thus training data, it can generate to facilitate online learning. Therefore many LTA systems also include an \emph{exploration} component which drives the gathering of new experiences. For example, CoBot robots were able to choose navigation routes which provided updated observations for environmental models~\cite{korein2014constrained}, and the STRANDS robots balanced exploration and exploitation to maximise interactions with humans during an information provision task~\cite{Hanheide2017} and to build 3D maps for object discovery~\cite{Ambrus2014,santos2016ral}.

Given the richness and diversity of techniques in 
machine learning,
many approaches 
could influence 
the ability of LTA systems to learn on the job in the future. Techniques which allow robots to continually learn from experience such as reinforcement learning,
or focus on particular experiences (e.g. failures, novelty) such as learning from demonstration %~\cite{Zeestraten17RAL} 
should allow online improvement of capabilities. Problems due to limited training in a particular domain (or open worlds) can be addressed by transfer learning, 
and supported by work from the exploration and active learning 
communities. 
Ongoing research is also investigating deep learning methods for long-term autonomy, including recent work on prediction of 
human trajectories from long-term observations~\cite{Sun2018ICRA}.

\section{FUTURE CHALLENGES}
\label{sec:challenges}

This paper discussed the LTA-related challenges in different areas of AI and
the importance of system-level integration 
for unlocking the potential of AI technologies.
In addition, we see the following major future challenges for LTA systems in real-world environments:

\paragraph*{Human-in-the-Loop Systems} 
How can LTA systems leverage human knowledge in unforeseen situations within long-term scenarios? As LTA systems have to deal with \emph{open worlds}, they will certainly require additional information when facing situations that were not foreseen at design time. This additional input might be given by end-users, maintainers, and/or domain experts. It might also be provided through direct control (i.e.\ teleoperation), natural interaction (e.g.\ via language or gestures) or 
labelled examples and/or data sets (e.g.\ via crowd-sourcing). To this end, LTA systems require mechanisms to integrate new, but potentially conflicting and/or untrustworthy, information in their KR about the world. This also requires that representations have some kind of semantic abstraction that can be linked to human knowledge. 
The fan-out~\cite{Olsen2004}, or number of robots a human can control simultaneously, will help drive the mixture of human supervisors and robots in such a paradigm.

\paragraph*{Knowledge Transfer between LTA Systems} What information about the world should be exchanged by LTA systems, and when? As more and more LTA robot systems get deployed, they can exchange important information to help  bootstrap other systems and/or to improve their performance.  As it is not realistic that all logged information is exchanged, it is important to investigate what information should be exchanged, and when. This also opens up privacy and security concerns. 
In this context, we believe that cloud-enabled knowledge bases \cite{riazuelo15roboearth} and other approaches in cloud robotics \cite{kehoe15cloudrobotics} will play an important role.

\paragraph*{Systems Integration}
Building robotic systems capable of long-term operations is inherently also a software engineering challenge, 
as they require the close integration of different AI techniques, 
a challenge also highlighted by~\cite{RAJAN20171}. While ROS has established itself as a de-facto standard framework for building integrated robotic systems, it provides only few instruments to ensure reliable and robust system architectures. Here, model-based approaches~\cite{Schlegel2015} might pave the way towards more dependable and verifiable integrated systems in the future.

\paragraph*{More Domain Specialisation}
Alongside the development of general principles of AI for long-term autonomy, there will be many interdependencies and synergies from solving the application-specific challenges in parallel.
For example, in precision agriculture
the accuracy of relative positioning and navigation, e.g.\ with respect to crop rows,
is more important than that of absolute navigation and position as provided by RTK GPS. 
Therefore, any improvements in recognition of crops
would in turn improve the robustness and accuracy of 
navigation in crop care and harvesting tasks.

\paragraph*{Verification and Evaluation of LTA Systems} 
How can the behaviour and the performance of LTA systems be verified and evaluated when robot system (including its models), task specification, and environment are constantly changing (at different timescales)? This requires LTA systems to keep a record of all their internal models that were used at a given time. Furthermore, it requires novel ways to provide formal  guarantees under the assumption that parts of the environments might change (with some probability)~\cite{DBLP:journals/corr/SeshiaS16}.

\paragraph*{Conclusion} 
Further to these technological challenges, we also see ethical, social, and legal issues when realising LTA systems, though these are beyond the scope of this paper. 
Overall, we believe strongly that AI methods can provide LTA systems with many of the capabilities needed to overcome these challenges.
In turn, rather than merely extending the lifetime of existing  AI-enabled robots, AI approaches may actually help to solve some of the really tough open problems in robotics, e.g.\ perception-based mobile manipulation in real-world settings, by leveraging long-term experience.
However, we recognise that, 
despite recent progress, 
there are still many exciting open challenges.

\bibliographystyle{IEEEtran}
\bibliography{survey}

\end{document}